\documentclass[sigplan,screen,nonacm]{acmart}
\AtBeginDocument{%
  }

\usepackage{multirow}
\usepackage{color}
\usepackage{tabularx}

\setcopyright{acmlicensed}
\copyrightyear{2026}
\acmYear{2018}
\acmDOI{XXXXXXX.XXXXXXX}
\acmConference[Conference acronym 'XX]{Make sure to enter the correct
  conference title from your rights confirmation email}{June 03--05,
  2018}{Woodstock, NY}
\acmISBN{978-1-4503-XXXX-X/2018/06}

\begin{document}

\title{LASA: A Weak Supervision Method for Open-Vocabulary Scene Sketch Semantic Segmentation}

\author{Liwen Yi}
\affiliation{%
  \institution{Beijing University of Posts and Telecommunications}
  \city{Beijing}
  \country{China}
}

\author{Xianlin Zhang}
\affiliation{%
  \institution{Beijing University of Posts and Telecommunications}
  \city{Beijing}
  \country{China}
}

\author{Yue Zhang}
\affiliation{%
  \institution{Beijing University of Posts and Telecommunications}
  \city{Beijing}
  \country{China}
}

\author{Yue Ming}
\affiliation{%
  \institution{Beijing University of Posts and Telecommunications}
  \city{Beijing}
  \country{China}
}

\author{Xueming Li}
\affiliation{%
  \institution{Beijing University of Posts and Telecommunications}
  \city{Beijing}
  \country{China}
}

\renewcommand{\shortauthors}{Trovato et al.}

\begin{abstract}
Open-vocabulary scene sketch semantic segmentation aims to assign dense semantic labels to sparse line drawings based on flexible category vocabularies specified at inference time, without relying on pixel-level annotations during training. Unlike natural images, sketches lack texture and color cues, making semantic understanding heavily dependent on stroke layout and spatial configuration, a challenge that renders single-layer vision-language features inherently unstable. Our key observation is that attention maps from different Vision Transformer layers encode complementary spatial cues: shallow layers capture global structural layouts, while deeper layers focus on local stroke intersections and object parts. This suggests that cross-layer aggregation provides a more robust structural prior than any individual layer alone. Leveraging this insight, we propose a structure-aware framework built upon \textbf{L}ayer-wise \textbf{A}ccumulated \textbf{S}tructural \textbf{A}ttention (\textbf{LASA}), which aggregates multi-layer attention to guide hierarchical semantic alignment under weak supervision and refine predictions during inference. Experiments on FS-COCO, SFSD, and FrISS show that LASA improves mIoU by $+3.43$, $+8.01$, and $+15.74$ over the prior weakly supervised baselines, demonstrating consistent gains in both segmentation accuracy and spatial coherence. Our source code will be made publicly available.

\end{abstract}

\ccsdesc[500]{Computing methodologies~Scene understanding}
\ccsdesc[500]{Computing methodologies~Image segmentation}

\keywords{Open-vocabulary, Scene Sketch, Semantic Understanding, Weak Supervision, Semantic Segmentation}


\maketitle
\begin{figure}[t]
  \centering
  \includegraphics[width=\linewidth]{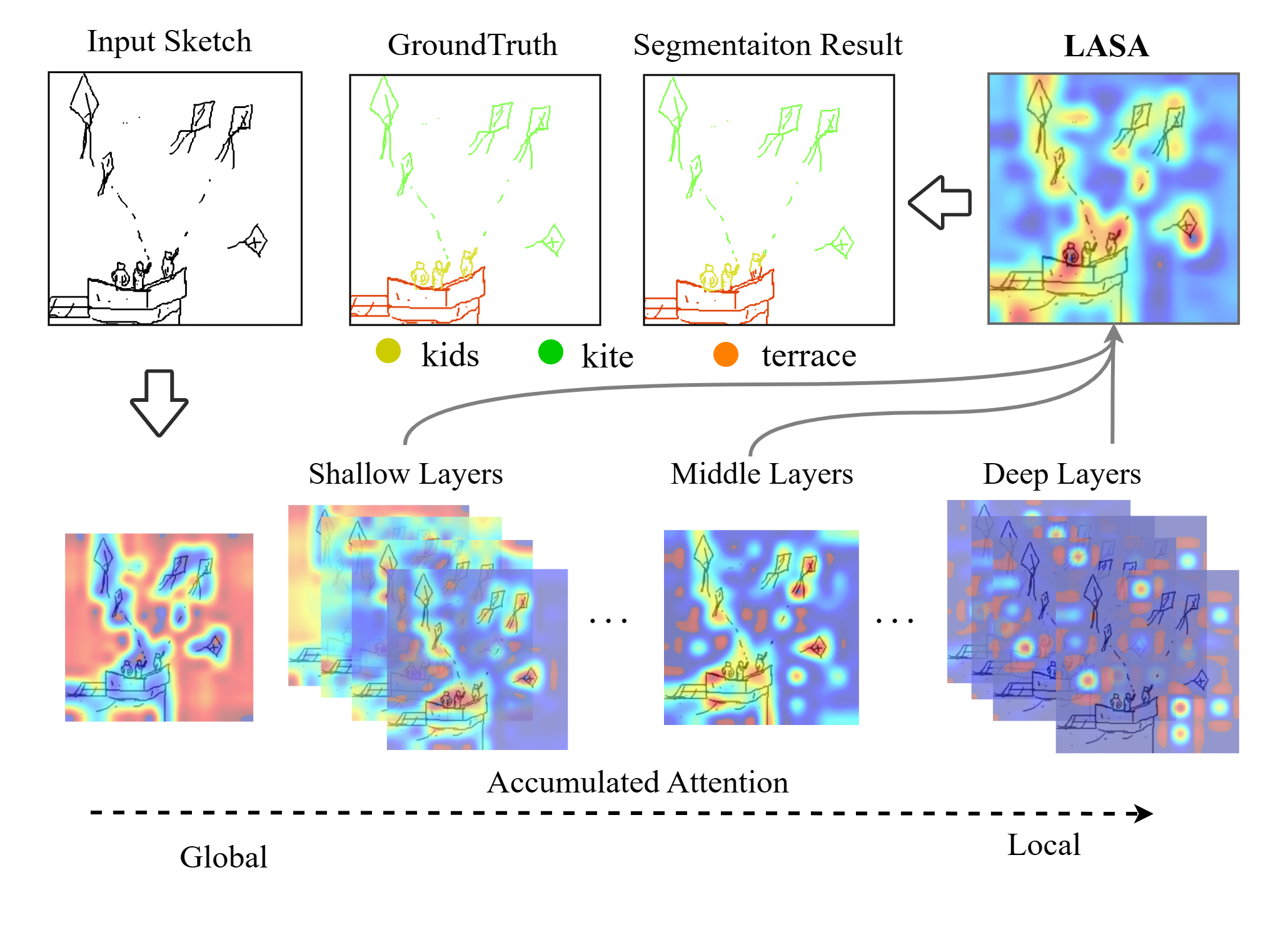}
  \caption{Illustration of the proposed LASA. It aggregates cross-layer structural attention to achieve more complete and
accurate open-vocabulary sketch segmentation, seamlessly fusing global layouts from shallow layers with fine-grained
stroke details from deeper ones.}
  \label{fig:teaser}
\end{figure}
\section{Introduction}
Open-vocabulary scene sketch semantic segmentation aims to assign dense semantic labels to sparse line drawings using category names provided only at inference time. This task addresses a critical need in human-centric creative workflows, where scene sketches serve as a universal medium for rapid ideation and design. However, acquiring dense pixel-level annotations for sketches is prohibitively expensive and does not scale, making weak supervision from image-level tags or captions an essential training paradigm~\cite{zou2018sketchyscene,chowdhury2022fscoco,bourouis2024ovsss}. Despite its practical importance, this task faces two fundamental challenges: \textit{(1) Domain mismatch in vision-language models.} Existing open-vocabulary segmentation pipelines are predominantly built on vision-language pretraining (e.g., CLIP~\cite{radford2021clip}) and optimized for appearance-rich natural images~\cite{rao2022denseclip,xu2022groupvit,xu2023san,liang2023ovseg}. When directly generalized to texture-free sketches, these models manifest substantial performance degradation: deep-layer features become excessively appearance-biased and spatially incoherent, consequently yielding fragmented regions, ill-defined boundaries, and insufficient object coverage.~\cite{lan2024proxyclip,sun2025cliper}. \textit{(2) Under-constrained spatial supervision.} Under weak supervision, learning signals primarily anchor global semantics while leaving region-level spatial cues poorly constrained. This limitation is particularly detrimental for sketches, where semantic information is encoded almost exclusively through stroke structure rather than appearance cues. The resulting structure-semantics mismatch renders robust localization especially challenging~\cite{lin2023clipes,he2023clips4,zhang2024frozenclip,bourouis2024ovsss}.

Through systematic analysis of Vision Transformer (ViT) behavior on sketch data, we make a key observation: attention maps across different ViT layers provide complementary structural evidence for scene sketches. Specifically, shallow layers capture broad scene layout and global composition, whereas deeper layers emphasize local stroke intersections and discriminative object parts (Fig.~\ref{fig:attention_layers}). This hierarchical structure suggests that aggregating cross-layer attention can yield a more complete and stable structural prior than relying on any single layer~\cite{abnar2020attentionflow,caron2021dino,wang2022tokencut,oquab2023dinov2}. Motivated by this observation, we propose \textbf{L}ayer-wise \textbf{A}ccumulated \textbf{S}tructural \textbf{A}ttention (\textbf{LASA}), as illustrated in Fig.~\ref{fig:teaser}, a structure-aware framework for open-vocabulary scene sketch segmentation. 

Our approach consists of three integrated components: \textbf{Accumulated  Attention Encoder:} aggregates multi-layer ViT attention into a unified structural prior and injects it into final-layer feature fusion, providing explicit structural guidance for texture-free sketch representations. \textbf{Hierarchical Semantic Alignment:} a weak supervision strategy that combines scene-level alignment for global semantic anchoring with LASA-guided region-aware alignment for structurally salient localization. \textbf{LASA-based Attention Refinement:} an inference-time module that fuses multi-layer prediction scores with LASA-based propagation to enhance spatial coherence and boundary precision. We conduct comprehensive evaluations on three established benchmarks: FS-COCO~\cite{chowdhury2022fscoco}, SFSD~\cite{zhang2023sfsd}, and FrISS~\cite{kutuk2025cavt}. Our results demonstrate consistent improvements over representative training-free and weakly supervised baselines across all datasets. Significant gains compared to the prior state-of-the-art weakly supervised method~\cite{bourouis2024ovsss}: $+3.43$ mIoU on FS-COCO, $+8.01$ mIoU on SFSD, and $+15.74$ mIoU on FrISS. Strong generalization to both pixel-level and stroke-level evaluation metrics, confirming the effectiveness of explicit cross-layer structural priors for sparse visual domains. To summarize, the contributions of this work are listed as follows:
\begin{itemize}
    \item  We reveal that different ViT layers encode complementary spatial cues for scene sketches, and consequently propose \textbf{LASA} to aggregate them into a stable cross-layer structural prior.

    \item We propose an \textbf{Accumulated Attention Encoder} capable of progressively aggregating attention features across varying depths of the network.
    
    \item  Building on LASA, this framework features \textbf{Hierarchical Semantic Alignment} to supervise the training process and an \textbf{LASA-based Attention Refinement} mechanism to optimize inference.
    
    \item We conduct extensive evaluations on three benchmarks, establishing new state-of-the-art results for weakly supervised open-vocabulary sketch segmentation with consistent improvements across the board.
\end{itemize}

\section{Related Work}

\subsection{Scene Sketch Segmentation}
Scene sketch segmentation has progressed from single-object tasks such as
recognition~\cite{eitz2012humansketch,yu2017sketchanet} and
retrieval~\cite{sangkloy2016sketchydb,yu2016sketchshoe,liu2017dsh,song2017fgsbir,dey2019doodle}
to sketch parsing and labeling~\cite{huang2014seglabel}, and then to multi-object scene understanding,
where structural relations and stroke context become the primary semantic
carriers~\cite{zou2018sketchyscene,zhang2018context,chowdhury2022fscoco,gao2020sketchycoco}.
Empowered by further developments, subsequent work improved local detail modeling and Transformer-based feature
fusion for fully supervised closed-set
settings~\cite{ge2022localdetail,zhang2023sfsd,yang2023sketchseger,zheng2023sketchsegformer},
and recent efforts have pushed toward instance- and stroke-level
understanding~\cite{kutuk2025cavt,tang2025inklayer}.
However, these methods either require dense pixel-level annotations or operate
over a fixed category set, making them unsuitable for the weakly supervised
open-vocabulary setting. In contrast, our method targets weakly supervised open-vocabulary scene sketch segmentation and introduces LASA to exploit cross-layer structural cues from CLIP itself, enabling stronger structural localization without dense masks or fixed vocabularies.

\subsection{Open-Vocabulary-based Semantic Segmentation }
In the realm of open-vocabulary semantic segmentation, the majority of existing works are tailored for natural images. These approaches can be broadly categorized into supervised and training-free methods. Supervised approaches~\cite{mukhoti2023pacl,zhou2023zegclip,ding2023maskclip,cho2024catseg,liu2024scan,shan2024ebseg} typically address this dense prediction task through pixel/patch alignment, e.g. DenseCLIP~\cite{rao2022denseclip}, LSeg~\cite{li2022lseg}, SED~\cite{xie2024sed} etc., or by employing grouping and mask-decoding pipelines ~\cite{xu2022groupvit,xu2023odise,qin2023freeseg,zhang2023openseed,zou2023xdecoder}, e.g. SAN~\cite{xu2023san}, OVSeg~\cite{liang2023ovseg}, OPSNet~\cite{chen2023opsnet} etc.. However, these paradigms intrinsically rely on annotated data and are restricted to fixed categories.
Conversely, training-free methods~\cite{zhang2025corrclip,sun2025cliper} pivot to injecting spatial priors from auxiliary foundation models. For instance, ProxyCLIP~\cite{lan2024proxyclip} utilizes DINO~\cite{caron2021dino} or SAM~\cite{kirillov2023segmentanything} features as proxy attention to reorganize CLIP's value embeddings. Related directions~\cite{yuan2024ovsam,stojnic2025lposs,lee2025escnet} include diffusion/SAM-assisted or propagation-based training-free designs such as FreeDA~\cite{barsellotti2024freeda}, PnP-OVSS~\cite{luo2024pnpovss}. While highly effective on natural images, these methods ~\cite{lin2023clipes,he2023clips4,zhang2024frozenclip,liu2025openbench,ulger2025autovocabulary} heavily depend on auxiliary models pre-trained on appearance-rich data. As scene sketches inherently lack texture, making structural layout their sole semantic carrier, these spatial priors cannot be reliably transferred to texture-free sketch domains.
The closest prior work to ours is introduced by Bourouis et al.~\cite{bourouis2024ovsss}, who proposed a weakly supervised open-vocabulary framework specifically for scene sketches. However, our approach distinguishes itself by deriving a robust structural prior entirely from CLIP's internal cross-layer attention, completely circumventing the need for any external auxiliary models.

\subsection{Structural Priors in ViT}
It is widely acknowledged that ViT attention maps encode implicit structural information whose utility extends far beyond mere content matching~\cite{dosovitskiy2021vit}. Specifically, mechanisms such as attention rollout and flow aggregate attention across layers to better capture token-level information~\cite{abnar2020attentionflow}. Concurrently, self-supervised ViTs have been shown to exhibit emergent structural cues highly applicable to object discovery and segmentation~\cite{caron2021dino,wang2022tokencut,oquab2023dinov2}. However, these insights are predominantly validated on natural images. Furthermore, they are typically employed either for post-hoc analysis or as standalone mask extraction tools, rather than being explicitly integrated as train-time structural priors within weakly supervised open-vocabulary sketch segmentation. In contrast, our method turns cross-layer attention aggregation into a task-specific structural prior for sparse scene sketches and integrates it into feature fusion, weakly supervised alignment, and inference-time refinement.

\begin{figure*}[t]
  \centering
  \includegraphics[width=\textwidth]{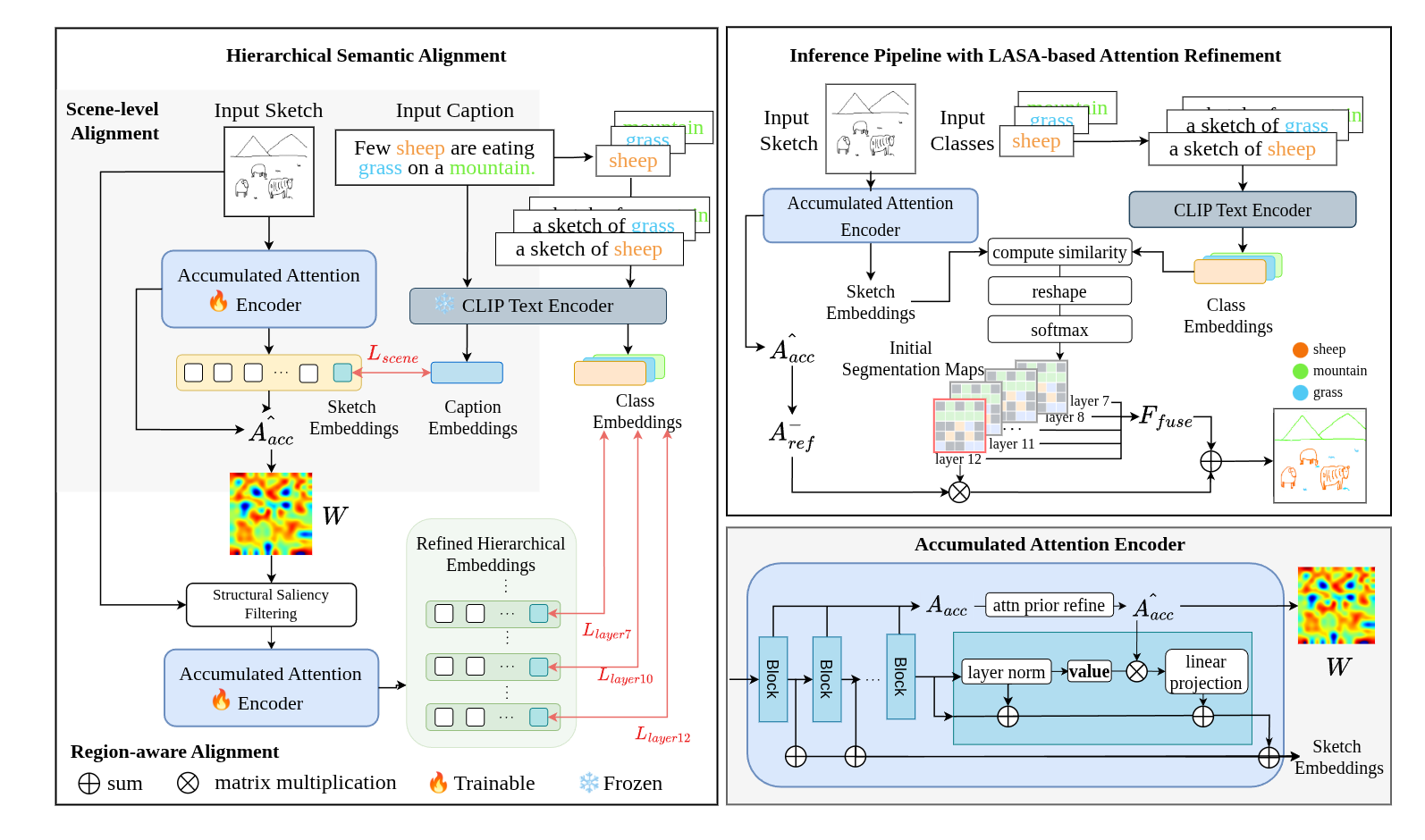}
  \caption{Overview of the proposed LASA. Given a scene sketch and textual category descriptions, multi-layer self-attention from the Accumulated Attention Encoder is accumulated to form layer-wise accumulated structural attention. Under weak supervision, a Hierarchical Semantic Alignment strategy is adopted: scene-level alignment matches the CLS feature with text embeddings to provide a global semantic anchor, while LASA-guided region-aware Alignment uses LASA-filtered re-encoding to emphasize structurally salient regions. At inference time, the class activation maps are dynamically refined by fusing the averaged multi-layer scores with the last-layer score map, which is propagated via the final LASA attention, through a residual addition. This combined process effectively mitigates noise and bolsters spatial coherence for open-vocabulary segmentation.}
  \Description{Framework diagram for the proposed open-vocabulary scene sketch segmentation method with LASA, hierarchical semantic alignment, and inference-time refinement.}
  \label{fig:framework}
\end{figure*}

\section{Method}
\subsection{Overview}
\textbf{Problem Formulation.} Formally, we define a scene sketch as an image $S \in \mathbb{R}^{H_{\text{img}} \times W_{\text{img}}}$ composed of binary or sparse line strokes, and a set of open-vocabulary semantic categories $\mathcal{C} = \{c_1, c_2, \dots, c_C\}$ where each category $c_i$ is described by a natural language phrase. The goal is to learn a segmentation function $f_\theta : (S, \mathcal{C}) \rightarrow M$ that predicts a dense semantic map $M \in \{1, \dots, C\}^{H_{\text{img}} \times W_{\text{img}}}$ assigning each pixel to one of the categories in $\mathcal{C}$. Different from closed-set segmentation, the category set $\mathcal{C}$ is not fixed during training and can be provided as flexible text prompts at inference time.

As illustrated in Fig.~\ref{fig:framework}, the proposed LASA mainly consists of three core components: accumulated  attention encoder for integrating cross-layer spatial cues, hierarchical semantic alignment for weakly supervised training, and LASA-based attention refinement for improving spatial coherence at inference time. Concretely, training phase: an input scene sketch $S$ is fed into the accumulated attention encoder (as detailed in Sec. 3.2) to extract self-attention maps 
$\{\mathbf{A}^{(l)}\}$ across different layers. Through accumulation and normalization, these maps are synthesized into a structural prior $\hat{\mathbf{A}}_{\mathrm{acc}}$, which is injected into the final feature aggregation layer to produce the sketch embeddings. For optimization, the encoded triplets first undergo a scene-level semantic alignment between the raw sketch features and the scene text prompts. Specifically, region-aware alignment is further conducted between the category text features and the sketch representations aggregated via the accumulated attention. Finally, the network is trained via joint optimization (as presented in Sec. 3.3). Inference phase: at inference time, taking $S$ and an open-vocabulary category set $\mathcal{C}$ as inputs, the model first predicts multi-layer category response maps $\{\mathbf{F}^{(l)}\}$
. These maps are subsequently refined through LASA-based attention refinement (as introduced in Sec. 3.4), yielding the final pixel-level scene sketch semantic segmentation results. In the following subsections, we elaborate on each key component of LASA.

\begin{figure}[t]
  \centering
  \includegraphics[width=\linewidth]{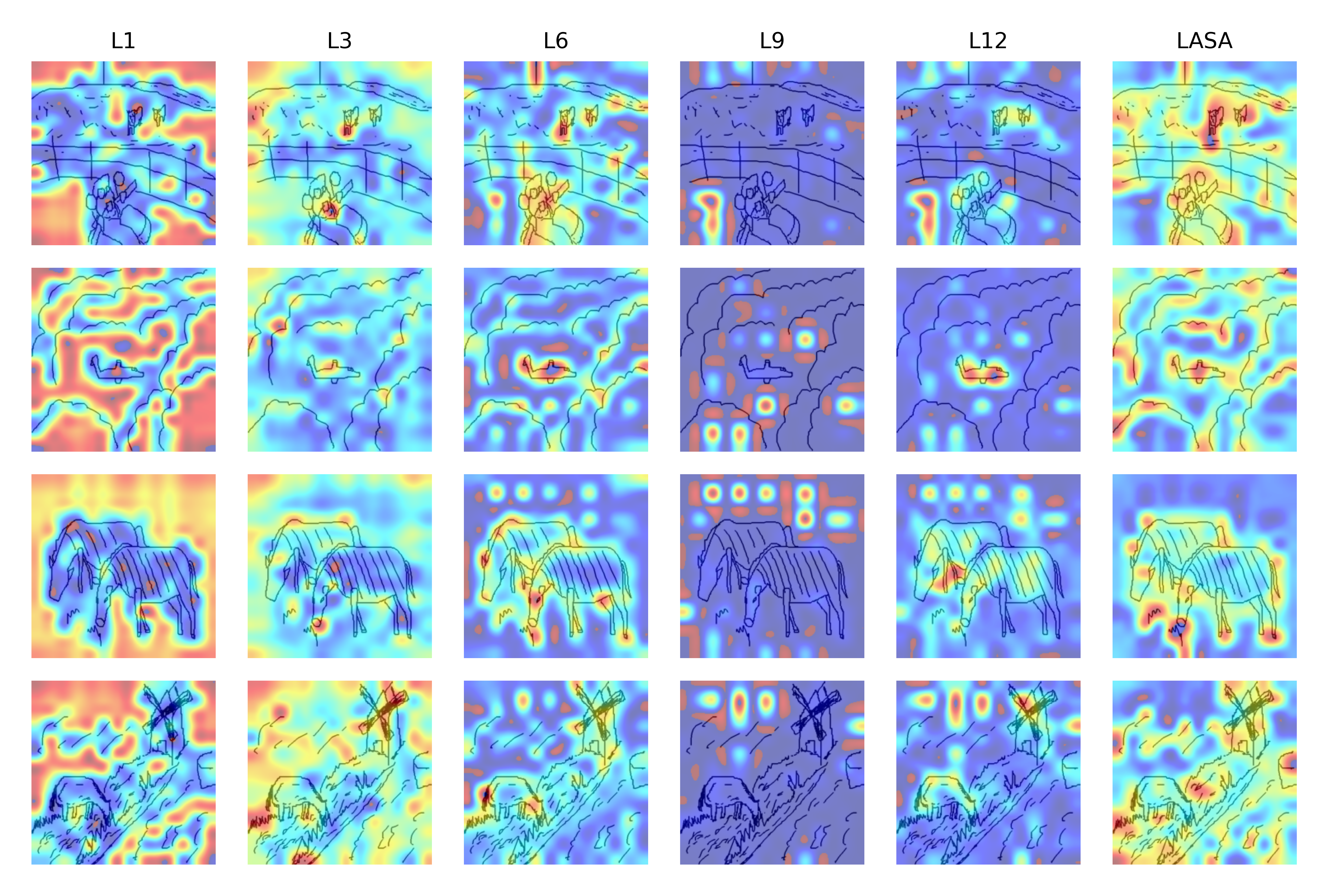}
  \caption{Attention maps from selected Transformer layers (L1, L3, L6, L9, L12) and the proposed LASA across four scene sketches. Shallow layers produce diffuse responses covering broad spatial regions, while deeper layers generate localized activations around stroke intersections and object parts. LASA integrates these complementary patterns into a unified structural representation, motivating its use as a spatial prior throughout the proposed framework.}
  \Description{Comparison of attention maps across several transformer layers and the proposed accumulated structural attention on scene sketch examples.}
  \label{fig:attention_layers}
\end{figure}

\subsection{Accumulated Attention Encoder}
\label{sec:lasa}
As illustrated in Fig.~\ref{fig:attention_layers}, attention maps from different Transformer layers exhibit complementary behaviors when processing scene sketches. Shallow layers tend to produce diffuse responses over broader regions, whereas deeper layers generate more localized attention peaks around stroke intersections or object parts. Since individual layers provide only partial spatial cues, aggregating them across layers can yield more complete token relations. Motivated by this observation, we design the accumulated attention encoder to learn and aggregate attention across different layers.

\textbf{Layer-wise Attention Accumulation.} Let $\mathbf{A}^{(l)} \in \mathbb{R}^{N_h \times N \times N} $ $\quad l = 1, \dots, L-1,$ denote the self-attention matrices extracted from the first $L-1$ Transformer layers, where $N_h$ is the number of attention heads. We first averages over the $N_h$ attention heads within each layer to obtain a per-layer relation matrix $\bar{\mathbf{A}}^{(l)} \in \mathbb{R}^{N \times N}$:
\begin{equation}
\bar{\mathbf{A}}^{(l)} = \frac{1}{N_h}\sum_{h=1}^{N_h} \mathbf{A}^{(l)}_h
\end{equation}
and then aggregates these matrices across layers:
\begin{equation}
\mathbf{A}_{\text{acc}} = \sum_{l=1}^{L-1} \bar{\mathbf{A}}^{(l)} \in \mathbb{R}^{N \times N}
\end{equation}
Compared with single-layer attention, the accumulated attention integrates complementary token relations from multiple layers and provides more complete spatial cues for sparse scene sketch understanding.

\textbf{Structure-prior Attention Fusion.} To form a valid spatial prior, the accumulated attention $\mathbf{A}_{\text{acc}}$ is first symmetrized by averaging with its transpose: $\mathbf{A}_{\text{sym}} = (\mathbf{A}_{\text{acc}} + \mathbf{A}_{\text{acc}}^\top) / 2$. Then, row-wise $\ell_1$ normalization is applied and negative values are removed by ReLU, yielding $\hat{\mathbf{A}}_{\text{acc}}$.
The accumulated attention is injected into the final Transformer layer as this layer produces the token representations directly consumed by downstream semantic alignment and inference, replacing its attention with the accumulated structural prior is therefore the most direct way to propagate structural guidance into the output features. Concretely,  layer-wise accumulated attention replaces standard content-based attention and guides feature aggregation as:
\begin{equation}
\mathbf{Y} = \hat{\mathbf{A}}_{\text{acc}} \mathbf{V},
\end{equation}
where $\mathbf{V} \in \mathbb{R}^{N \times D}$ denotes the concatenated value features across all heads and $\mathbf{Y} \in \mathbb{R}^{N \times D}$ is the resulting token representation.
This formulation differs from standard multi-head attention, where each head maintains its own attention map before concatenation. Instead, our implementation can be viewed as a head-collapsed structural approximation: the accumulated attention is reduced to a shared structural prior, broadcast to all heads, and then used for head-wise value aggregation in the final layer. The trade-off is that head-specific routing is not preserved explicitly, in return, the model receives a unified cross-layer structural prior that is better aligned with our goal of emphasizing stable sketch relations over head-dependent content matching. As a result, this design transforms attention from single-layer content matching into cross-layer spatially guided feature fusion, which can help the resulting visual representations better preserve sketch layouts and stroke groupings.

\subsection{Hierarchical Semantic Alignment}
\label{sec:hierarchical_alignment}
To train the proposed framework without pixel-level annotations, we introduce a hierarchical semantic alignment strategy, which integrates scene-level alignment and region-aware alignment under the guidance of LASA. Specifically, this strategy comprises a dual-alignment mechanism: scene-level semantic alignment and region-aware alignment.

\textbf{Scene-level Semantic Alignment.} Given a mini-batch of paired sketch-text samples $\{(S_i, t_i, y_i)\}_{i=1}^{B}$, where $t_i$ is the scene description paired with sketch $S_i$ and $y_i$ denotes the scene label of the $i$-th sketch-caption pair, we optimize a batch-wise triplet loss with hardest in-batch negative mining:
\begin{equation}
\begin{aligned}
\mathcal{L}_{\text{scene}}
&= \frac{1}{B}\sum_{i=1}^{B}\max(
0,\,
\delta + d_{1}(f_{\text{cls}}(S_i), g(t_i)) \\
&\quad - \min_{j:\,y_j \neq y_i}
d_{1}(f_{\text{cls}}(S_i), g(t_j)))
\end{aligned}
\end{equation}
where $f_{\text{cls}}(S_i)$ denotes the visual feature of the classification token (CLS) extracted from sketch $S_i$, $d_{1}(\cdot,\cdot)$ denotes the $\ell_{1}$ distance, and the negative term is selected as the minimum-distance text embedding among samples with labels different from $y_i$ within the current mini-batch.
This global alignment provides a scene-level semantic anchor and can help reduce excessive reliance on noisy local regions during training.

\textbf{Region-aware Alignment.} 
A remaining challenge after scene-level alignment is that the global CLS supervision alone does not explicitly emphasize which spatial regions should dominate the visual representation. To inject spatially localized cues into semantic learning while preserving spatial coherence, we introduce region-aware alignment, which is implemented through LASA-filtered re-encoding. First,
a spatial saliency map $\mathbf{W}$ is derived from the fused attention by extracting the CLS-to-patch responses, reshaping them into a spatial grid, and upsampling the result to the input resolution. As a result, $\mathbf{W} \in \mathbb{R}^{H_{\text{img}} \times W_{\text{img}}}$ assigns each image location a structural saliency score that reflects how strongly it is attended by the global CLS token under LASA guidance. To suppress noisy responses, we apply a fixed threshold $\tau_c$ and retain only high-confidence locations by hard thresholding:
\begin{equation}
\tilde{\mathbf{W}} = \mathbb{I}(\mathbf{W} > \tau_c) \odot \mathbf{W},
\end{equation}
where $\mathbb{I}(\cdot)$ is the indicator function and $\tau_c$ is treated as a fixed hyper-parameter during training only.
The filtered saliency map $\tilde{\mathbf{W}}$ is then used to construct the LASA-filtered re-encoded input. Let $\mathbf{B} = 1 - \mathbf{S}$ denote the stroke mask of the sketch image. We first compute
\begin{equation}
\mathbf{Q} = \tilde{\mathbf{W}} \odot \mathbf{B},
\end{equation}
and convert it back to the white-background style expected by CLIP via global max normalization:
\begin{equation}
\mathbf{S}_w = m_{\max} - \mathbf{Q}, \qquad
m_{\max} = \max_{u,v} \mathbf{Q}_{u,v},
\end{equation}
where the maximization is taken over all spatial locations $(u,v)$ in the 2D map. In this way, $\mathbf{S}_w$ preserves the original sketch format while emphasizing structurally salient regions identified by LASA.
Finally, $\mathbf{S}_w$ is re-encoded by the visual encoder. Since the re-encoded input is weighted by the LASA-derived saliency map, the resulting CLS representation is biased toward structurally salient regions rather than the entire sketch uniformly. This LASA-filtered re-encoding step introduces region-aware cues without introducing explicit region tokens or pixel-level supervision.
The re-encoded features are supervised with the same batch-wise triplet formulation used in scene-level alignment. For a selected layer set $\mathcal{I}$, LASA-guided region-aware alignment is defined on the CLS features of the LASA-filtered re-encoded sketch as:
\begin{equation}
\begin{aligned}
\mathcal{L}_{\text{region}}
&= \frac{1}{|\mathcal{I}|B}\sum_{l \in \mathcal{I}} \sum_{i=1}^{B}\max(
0,\,
\delta + d_{1}(f^{(l)}_{\text{cls}}(S_{w,i}), g(c_i)) \\
&\quad - \min_{j:\,y_j \neq y_i}
d_{1}(f^{(l)}_{\text{cls}}(S_{w,i}), g(c_j)))
\end{aligned}
\label{eq:region_loss}
\end{equation}

where $c_i$ denotes the category text paired with the $i$-th re-encoded sketch in the current mini-batch, and $y_i$ denotes its category label for in-batch negative mining. The negative term is mined from category embeddings whose labels differ from $y_i$. Here, $f^{(l)}_{\text{cls}}(S_{w,i})$ is the CLS feature extracted at layer $l$, $g(\cdot)$ is the text encoder, $d_{1}(\cdot,\cdot)$ is the $\ell_{1}$ distance, and $\delta$ is the triplet margin.
By enforcing semantic consistency across layers on LASA-weighted inputs, LASA-guided region-aware alignment is intended to encourage region-aware semantic cues that better follow sketch layouts and stroke groupings. At the same time, it remains fully compatible with the weakly supervised setting because the supervision still operates through paired text descriptions rather than pixel-level annotations.

The final training objective jointly optimizes scene-level alignment and region-aware alignment:
\begin{equation}
\mathcal{L} = \mathcal{L}_{\text{scene}} + \mathcal{L}_{\text{region}}
\end{equation}
This joint optimization strategy combines global semantic consistency with region-aware alignment based on cross-layer spatial cues, which is expected to improve open-vocabulary segmentation under weak supervision.

\subsection{LASA-based Attention Refinement}
\label{sec:refinement}
At inference time, class activation maps are obtained by computing the similarity between patch-level visual features and text embeddings under the guidance of LASA. Let $\mathbf{F}^{(l)} \in \mathbb{R}^{C \times N_p}$ denote the layer-wise class activation map at Transformer layer $l$ after flattening the spatial grid, where $N_p$ is the number of patch tokens and $C$ is the number of categories. Although LASA already injects accumulated structural priors into encoder features, predictions from a single layer can remain locally fragmented. The LASA-based refinement follows the implementation: (1) average selected layer-wise scores, (2) propagate the last-layer score map with the final-layer LASA attention, and (3) combine them by residual addition. For a selected layer set $\mathcal{I}_r$, we define the fused score map as:
\begin{equation}
\mathbf{F}_{\text{fuse}} = \frac{1}{|\mathcal{I}_r|}\sum_{l \in \mathcal{I}_r}\mathbf{F}^{(l)},
\label{eq:refine_fuse}
\end{equation}
The refinement attention matrix is given by the accumulated structural attention used in the final layer:
\begin{equation}
\mathbf{A}_{\text{ref}} = \hat{\mathbf{A}}_{\text{acc}}.
\label{eq:refine_attn}
\end{equation}
The CLS token is removed to obtain $\mathbf{A}_{\text{ref}}^{-} \in \mathbb{R}^{N_p \times N_p}$, and propagate the last-layer map:
\begin{equation}
\widetilde{\mathbf{F}}_{\text{last}} = \mathbf{F}^{(L)} \mathbf{A}_{\text{ref}}^{-},
\label{eq:refine_last}
\end{equation}
which corresponds to multiplying the flattened last-layer score map by the attention matrix in our implementation. The final refined map is
\begin{equation}
\mathbf{F}' = \widetilde{\mathbf{F}}_{\text{last}} + \mathbf{F}_{\text{fuse}},
\label{eq:refine_propagation}
\end{equation}
where $\mathbf{F}'$ denotes the refined class activation map. In our default setting, $\mathcal{I}_r$ for score fusion contains the last $6$ Transformer layers (validated in Section~\ref{sec:ablation}). For pixel-level prediction, $\widetilde{\mathbf{F}}_{\text{last}}$ and $\mathbf{F}_{\text{fuse}}$ are reshaped back to the token grid and then combined.
Compared to relying on a single-layer prediction, this residual refinement strategy elegantly couples the structure-aware propagation from the LASA prior with multi-layer score averaging. Consequently, it significantly improves spatial region coherence across object layouts and stroke groupings, and effectively mitigates the emergence of isolated noisy activations. Importantly, operating strictly as a post-hoc inference mechanism, it guarantees enhanced segmentation quality while introducing absolutely no supplementary training overhead.

\section{Experiments}

\begin{table*}[t]
  \centering
  \small
  \caption{Performance comparison (\%) on FS-COCO, SFSD, and FrISS datasets. Fine-tuned baselines are re-trained under the same weakly supervised protocol as ours, using the same CLIP ViT-B/16 backbone, loss function, and training hyperparameters.}
  \label{tab:main_results}
  \begin{tabular}{c l ccc ccc ccc}
    \toprule
    \multirow{2}{*}{} & \multirow{2}{*}{Method}
    & \multicolumn{3}{c}{FS-COCO}
    & \multicolumn{3}{c}{SFSD}
    & \multicolumn{3}{c}{FrISS} \\
    \cmidrule(lr){3-5}
    \cmidrule(lr){6-8}
    \cmidrule(lr){9-11}
    &
    & Acc@P $\uparrow$ & Acc@S $\uparrow$ & mIoU $\uparrow$
    & Acc@P & Acc@S & mIoU
    & Acc@P & Acc@S & mIoU \\
    \midrule
    \multirow{4}{*}{Training free}
    & CLIP~\cite{radford2021clip}
    & 24.92 & 25.25 & 12.33
    & 19.41 & 15.59 & 11.19
    & 42.57 & 43.19 & 21.03 \\
    & ProxyCLIP~\cite{lan2024proxyclip}
    & 79.48 & 80.84 & 61.60
    & 72.13 & 59.18 & 47.70
    & 75.46 & 67.00 & 65.00 \\
    & CLIPer~\cite{sun2025cliper}
    & 69.72 & 77.55 & 58.36
    & 65.07 & 51.77 & 43.37  
    & 75.87 & 70.14 & 67.82\\
    & CorrCLIP~\cite{zhang2025corrclip}
    & 63.66 & 65.35 & 41.74
    & 58.03 & 48.97 & 34.83
    & 69.90 & 65.43 & 58.94 \\
    \midrule
    \multirow{4}{*}{Fine-tuned}
    & CLIP~\cite{radford2021clip}
    & 31.80 & 31.51 & 14.03
    & 24.54 & 22.35 & 17.69
    & 54.66 & 57.78 & 28.01 \\
    & ProxyCLIP~\cite{lan2024proxyclip}
    & 79.77 & 81.23 & 61.65
    & 66.25 & 56.12 & 42.80
    & 76.01 & 68.55 & 65.49 \\
    & CLIPer~\cite{sun2025cliper}
    & 72.92 & 79.20 & 60.29
    & 72.78 & 56.77 & 48.29  
    & 79.81 & 71.93 & 71.39 \\
    & CorrCLIP~\cite{zhang2025corrclip}
    & 69.68 & 71.31 & 46.26
    & 64.57 & 59.63 & 51.27
    & 72.30 & 71.01 & 63.99 \\
    \midrule
    \multirow{1}{*}{Fully supervised}
    & SketchSeger~\cite{yang2023sketchseger}
    & -- & -- & --
    & 72.58 & 72.16 & 51.81
    & 38.87 & 34.78 & 17.35 \\
    \midrule
    \multirow{2}{*}{Weakly supervised}
    & Bourouis~\cite{bourouis2024ovsss}
    & 80.73 & 82.45 & 67.05
    & 74.74 & 70.60 & 52.94
    & 67.34 & 59.92 & 58.23   \\
    & \textbf{Ours}
    & \textbf{83.88} & \textbf{85.25} & \textbf{70.48}
    & \textbf{78.32} & \textbf{75.77} & \textbf{60.95}
    & \textbf{82.24} & \textbf{73.87} & \textbf{73.97} \\
    \bottomrule
  \end{tabular}
\end{table*}

\subsection{Experimental Setups}
\textbf{Dataset.} We evaluate our method on three sketch segmentation datasets: FS-COCO~\cite{chowdhury2022fscoco}, SFSD~\cite{zhang2023sfsd}, and FrISS~\cite{kutuk2025cavt}. FSCOCO-seg is an open-vocabulary scene sketch segmentation benchmark derived from COCO~\cite{chowdhury2022fscoco,bourouis2024ovsss}. It contains multi-object scene sketches with diverse categories and is designed for evaluating open-vocabulary segmentation under weak supervision. Because FSCOCO-seg does not provide a closed-set fully supervised training protocol, fully supervised sketch segmentation methods are not directly comparable, and experiments on this dataset mainly focus on open-vocabulary and weakly supervised approaches. SFSD and FrISS are classical sketch semantic segmentation datasets with dense pixel-level annotations. On these datasets, we compare our method with both open-vocabulary methods and fully supervised sketch segmentation approaches to analyze the performance gap under different supervision regimes. Although SFSD and FrISS provide dense pixel-level annotations, we do not use these annotations for training in our weakly supervised setting; they are reserved only for evaluation. In our experiments, the open-vocabulary property comes from predicting segmentation masks by matching visual features with text prompts rather than a fixed closed-set classifier. We note that the evaluated categories on SFSD and FrISS are still drawn from the benchmark label set, so this protocol mainly reflects open-vocabulary compatible weakly supervised segmentation.

\textbf{Metrics.} We report mean Intersection-over-Union (\textbf{mIoU}), Pixel Accuracy (\textbf{Acc@P}), and Stroke Accuracy (\textbf{Acc@S}) as evaluation metrics. mIoU evaluates region-level segmentation quality and is the main metric for comparison. Pixel Accuracy measures overall prediction correctness across pixels. Stroke Accuracy evaluates segmentation consistency at the stroke level. For datasets with vector stroke annotations, each stroke is assigned the label predicted for the majority of its constituent pixels. For SFSD and FrISS, which do not provide vector stroke annotations, we approximate strokes using connected components extracted from the raster sketch, and compute Stroke Accuracy by assigning each connected component the majority predicted label among its pixels.

\textbf{Implementation Details.}

We use CLIP ViT-B/16~\cite{radford2021clip} with a frozen text encoder and a fine-tuned visual encoder, train with AdamW ($1\times10^{-6}$) for 20 epochs on a single RTX 3090, and resize sketches to $224\times224$. Weak supervision uses image-level category tags and scene captions only: scene-level alignment applies a batch-wise triplet loss on CLS features of original sketches, while region-level alignment applies the same objective to LASA-filtered re-encoded sketches at layers 7/10/12, with hardest in-batch negatives and margin 0.2. Ablations on different supervision-layer choices are provided in the supplementary material. For SFSD/FrISS, captions are templated from category tags, and $\tau_c$ is fixed to 0.7 during training only. For fair fine-tuned baseline comparison, CLIP, ProxyCLIP, CLIPer, and CorrCLIP follow the same weak-supervision protocol and optimizer schedule.

\textbf{Baseline Methods.}
To comprehensively evaluate the proposed method, we compare LASA with four categories of baselines including: \textbf{(1) Training-free methods}: CLIP~\cite{radford2021clip}, ProxyCLIP~\cite{lan2024proxyclip}, CLIPer~\cite{sun2025cliper}, CorrCLIP~\cite{zhang2025corrclip}; \textbf{(2) Fine-tuned CLIP-based baselines under a unified weakly supervised protocol}: CLIP~\cite{radford2021clip}, ProxyCLIP~\cite{lan2024proxyclip}, CLIPer~\cite{sun2025cliper}, CorrCLIP~\cite{zhang2025corrclip}; \textbf{(3) Fully supervised sketch segmentation methods}: SketchSeger~\cite{yang2023sketchseger}; \textbf{(4) Weakly supervised open-vocabulary methods}: Bourouis~\cite{bourouis2024ovsss}.
For comparability, CLIP, ProxyCLIP, CLIPer, CorrCLIP, and Bourouis are re-trained under the same weakly supervised setup as ours, while SketchSeger is reported as a fully supervised reference on SFSD/FrISS.

\begin{figure*}[!t]
  \centering
  \includegraphics[width=\textwidth]{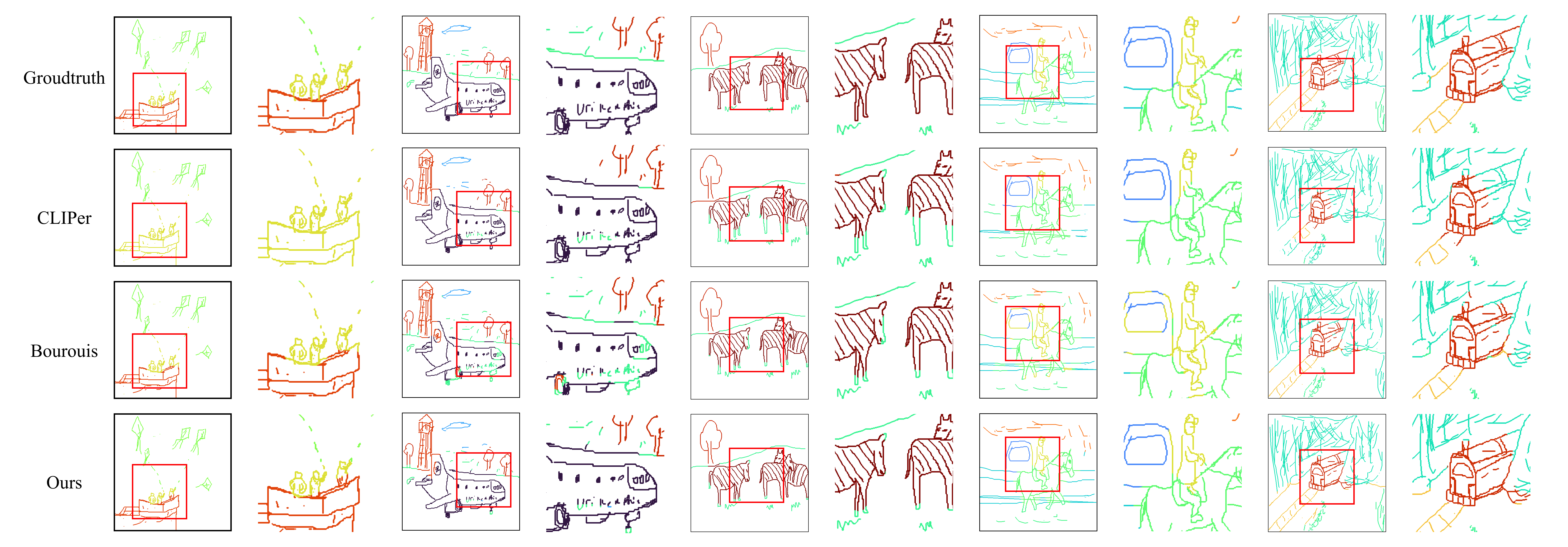}
  \caption{Qualitative comparison on representative scene sketches. From top to bottom: ground truth, CLIPer~\cite{sun2025cliper}, Bourouis~\cite{bourouis2024ovsss}, and ours. The even-numbered columns show localized enlargements of the areas highlighted by red boxes.}
  \Description{Qualitative comparison figure showing ground truth, CLIPer, Bourouis, and the proposed method on multiple scene sketch examples.}
  \label{fig:qualitative}
\end{figure*}

\subsection{Quantitative Results}

Table~\ref{tab:main_results} shows that LASA achieves the strongest weakly supervised open-vocabulary performance across all three benchmarks. Compared with Bourouis~\cite{bourouis2024ovsss}, LASA improves mIoU from 67.05 to 70.48 on FS-COCO (+3.43), from 52.94 to 60.95 on SFSD (+8.01), and from 58.23 to 73.97 on FrISS (+15.74). The same trend is reflected in Acc@P and Acc@S, indicating that the gains are consistent in both region-level and stroke-level evaluation.
Compared with the strongest training-free baselines, our method improves mIoU by +8.88 on FS-COCO (70.48 vs. 61.60, ProxyCLIP), +13.25 on SFSD (60.95 vs. 47.70, ProxyCLIP), and +6.15 on FrISS (73.97 vs. 67.82, CLIPer). Under the unified fine-tuning protocol, the strongest baseline is ProxyCLIP on FS-COCO (61.65), CorrCLIP on SFSD (51.27), and CLIPer on FrISS (71.39), while our method remains best on all three datasets, yielding +8.83, +9.68, and +2.58 mIoU improvements, respectively. These results suggest that LASA provides complementary structural benefits beyond optimization alone. We also report fully supervised SketchSeger~\cite{yang2023sketchseger} on SFSD/FrISS as a closed-set reference. Its substantially lower results under prompt-based evaluation (51.81 on SFSD and 17.35 on FrISS) suggest that fixed-category supervision may transfer less effectively to open-vocabulary inference. At the same time, the absolute gap may also be influenced by protocol mismatch between closed-set fully supervised training and our weakly supervised open-vocabulary evaluation. Overall, the quantitative evidence supports our central claim that explicit cross-layer structural priors improve weakly supervised open-vocabulary scene sketch segmentation. 

\subsection{Qualitative Results}
Fig.~\ref{fig:qualitative} demonstrates that LASA yields consistently more
spatially coherent masks than both CLIPer~\cite{sun2025cliper} and
Bourouis~\cite{bourouis2024ovsss} across FS-COCO, SFSD, and FrISS, with the
clearest gains in structurally complex scenes. When dealing with sophisticated open-vocabulary scene sketches, CLIPer often produces fragmented
regions and unstable boundaries around stroke intersections and overlapping
objects. Bourouis alleviates part of this fragmentation but still tends to
merge adjacent regions and miss thin boundaries in cluttered layouts. In
contrast, our method more consistently recovers complete object regions with
cleaner boundaries, suggesting that cross-layer structural priors preserve
stroke structure more effectively under weak supervision. This trend is
consistent across diverse scene compositions. Additional
qualitative visualizations are provided in the supplementary material. 

\subsection{Ablation Study}
\label{sec:ablation}
We conduct ablation experiments on FS-COCO to quantify the contribution of each component under the same weakly supervised protocol as the full model. The ablations are designed to isolate three factors: (1) cross-layer accumulated structural attention encoder, (2) hierarchical semantic alignment, and (3) inference-time attention refinement. Unless otherwise specified, all other settings are kept unchanged.

\textbf{Accumulated Attention  Encoder.}

\begin{figure}[t]
  \centering
  \includegraphics[width=\linewidth]{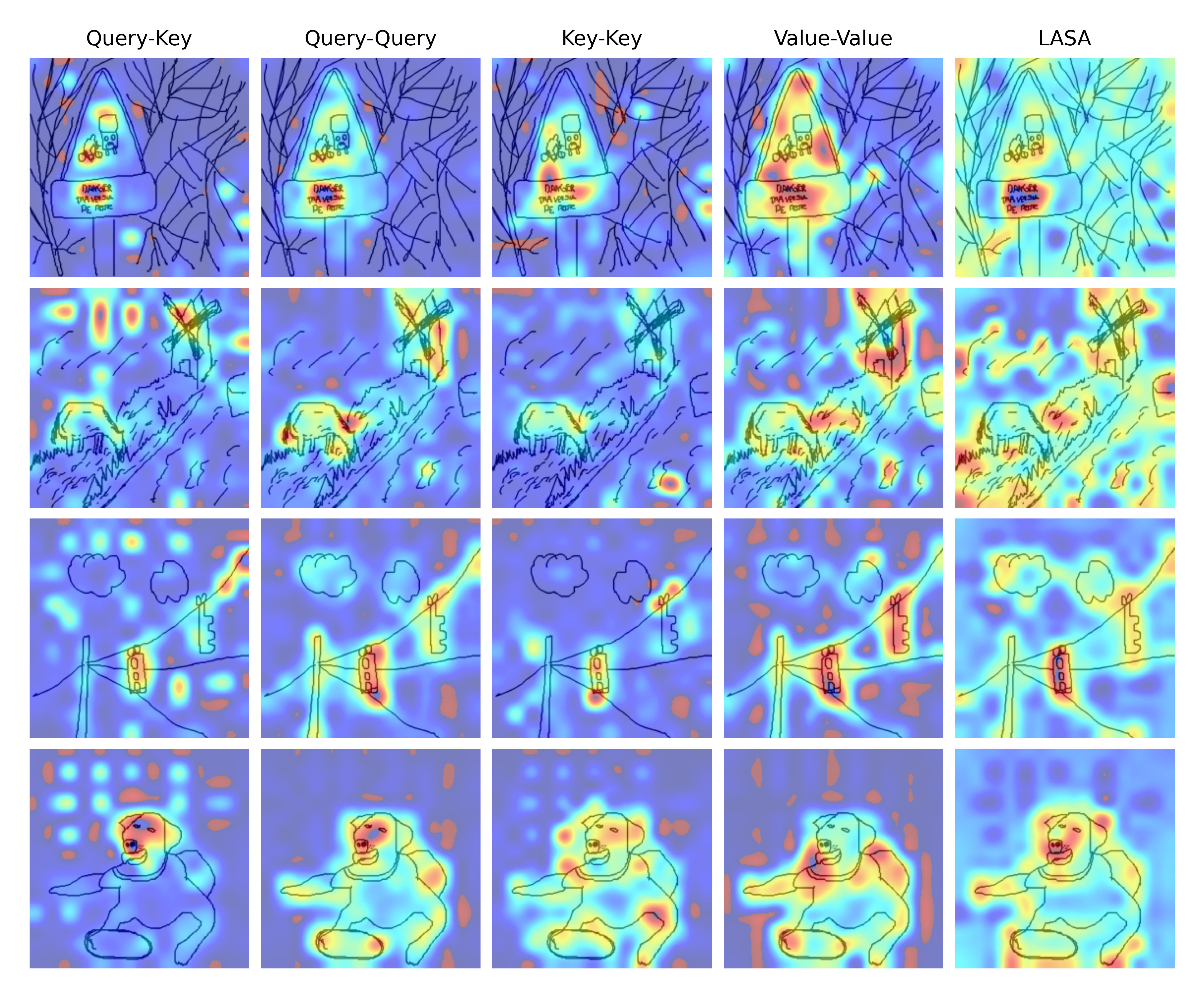}
  \caption{Attention formulation comparison (Query--Key, Query--Query, Key--Key, Value--Value, LASA). LASA produces the most coherent responses.}
  \Description{Visualization comparing several attention formulations and the proposed LASA on multiple scene sketches.}
  \label{fig:attention_ablation}
\end{figure}
To verify the effect of LASA, we replace cross-layer accumulated attention with single-layer alternatives while keeping the rest of the pipeline unchanged. Table~\ref{tab:ablation_all} (Attention formulation block) shows that LASA outperforms all formulations compared. In particular, relative to standard Query--Key attention, LASA improves Acc@P from 46.46 to 83.88, Acc@S from 46.06 to 85.25, and mIoU from 21.77 to 70.48. Even compared with the strongest non-LASA variant (Value--Value, 67.49 mIoU), LASA still gains +2.99 mIoU.
Fig.~\ref{fig:attention_ablation} provides a qualitative explanation for this gap: non-accumulated alternatives tend to produce fragmented or locally noisy responses, while LASA yields smoother and more object-aligned activation patterns along stroke structures. These observations are consistent with our hypothesis that cross-layer aggregation provides a more stable structural prior than relying on a single-layer attention pattern.

\begin{table}[t]
  \centering
  \small
  \setlength{\tabcolsep}{3pt}
  \caption{Unified ablation results on FS-COCO. For all metrics, the superior results are denoted in bold.}
  \label{tab:ablation_all}
  \begin{tabularx}{\columnwidth}{@{}>{\raggedright\arraybackslash}p{0.23\columnwidth} >{\raggedright\arraybackslash}X c c c@{}}
    \toprule
    Category & Setting & Acc@P & Acc@S & mIoU \\
    \midrule
    \multirow{5}{*}{\shortstack[l]{Attention\\formulation}}
    & Query-key & 46.46 & 46.06 & 21.77 \\
    & Query-query & 70.22 & 74.53 & 54.76 \\
    & Key-key & 65.51 & 76.47 & 56.28 \\
    & Value-value & 80.65 & 83.88 & 67.49 \\
    & \textbf{LASA} & \textbf{83.88} & \textbf{85.25} & \textbf{70.48} \\
    \midrule
    \multirow{6}{*}{\shortstack[l]{Component \\ ablation}}
    & w/o LASA & 46.46 & 46.06 & 21.77 \\
    & w/o loss & 74.42 & 81.04 & 62.94 \\
    & w/o global loss & 81.00 & 83.06 & 66.11 \\
    & w/o region-aware & 82.79 & 83.85 & 68.90 \\
    & w/o refinement & 69.65 & 82.52 & 66.67 \\
    & \textbf{full} & \textbf{83.88} & \textbf{85.25} & \textbf{70.48} \\
    \midrule
    \multirow{4}{*}{\shortstack[l]{Refinement \\ layers}}
    & without refinement & 69.65 & 82.52 & 66.67 \\
    & all layers & 83.00 & 84.38 & 68.89 \\
    & last 3 layers & 83.82 & 85.09 & 70.29 \\
    & \textbf{best (last 6 layers)} & \textbf{83.88} & \textbf{85.25} & \textbf{70.48} \\
    \bottomrule
  \end{tabularx}
\end{table}

\textbf{Hierarchical Semantic Alignment.}
To evaluate the contribution of each alignment level, we compare three ablated variants: removing both alignment losses (\textit{w/o loss}), removing only scene-level alignment (\textit{w/o global loss}), and removing only LASA-guided region-aware alignment (\textit{w/o region-aware}).
Table~\ref{tab:ablation_all} (Component ablation block) shows that removing both losses causes the largest degradation (mIoU 70.48$\rightarrow$62.94, a 7.54-point drop), confirming the necessity of hierarchical supervision in the weakly supervised setting. Removing only scene-level alignment reduces mIoU to 66.11 (4.37-point drop), while removing only LASA-guided region-aware alignment reduces mIoU to 68.90 (1.58-point drop). The same trend is reflected in Acc@P and Acc@S, indicating that the effect is not limited to one metric.
These results suggest complementary roles for the two levels: scene-level alignment acts as the primary semantic anchor for category discrimination, while region-aware alignment contributes additional region-level consistency on structurally salient areas. Together, they provide stronger supervision than either level alone.

\textbf{LASA-based Attention Refinement.}
We evaluate LASA-based attention refinement by varying the fusion layer set $\mathcal{I}_r$ (none, all 12, last 3, last 6; Table~\ref{tab:ablation_all}).All refinement variants improve over the no-refinement baseline. The best setting (last 6 layers) increases mIoU from 66.67 to 70.48 (+3.81), Acc@P from 69.65 to 83.88 (+14.23), and Acc@S from 82.52 to 85.25 (+2.73). This indicates that refinement improves both region-level quality and stroke-level consistency.Using all 12 layers yields weaker results (68.89 mIoU) than using the last 3 or last 6 layers, suggesting that including shallow-layer attention introduces over-smoothing during propagation. The last 6 layers provide the best balance between structural context and localization, and is as the default setting.

\section{Conclusion}
This paper introduces LASA, which tackles the challenges of weakly supervised open-vocabulary scene sketch segmentation. By leveraging the observation that different Transformer layers provide complementary spatial information, our Accumulated Attention Encoder seamlessly integrates multi-scale attention to bridge the gap between shallow and deep features. Combined with a Hierarchical Semantic Alignment strategy and inference-stage optimization, our framework significantly improves spatial coherence and semantic accuracy. The superior results achieved on various datasets underscore the efficacy of LASA as a robust solution for sparse, scene-level sketch understanding, setting a new benchmark for future research in this field.
\clearpage
\bibliographystyle{ACM-Reference-Format}
\bibliography{refs}

\end{document}